\documentclass[runningheads]{llncs}
\usepackage{microtype}
\usepackage{graphicx}
\usepackage{subfigure}
\usepackage{booktabs} 
\usepackage{hyperref}
\usepackage{amsmath}
\usepackage{amssymb}
\usepackage{algorithm}
\usepackage{algorithmic}
\usepackage{bbm}
\usepackage{times}
\usepackage{soul}
\usepackage{multirow}
\usepackage{graphicx}
\begin{document}
\title{Universal Distributional Decision-based Black-box Adversarial Attack with Reinforcement Learning}
%
%
\author{Yiran Huang\inst{1} \and
Yexu Zhou\inst{1} \and
Michael Hefenbrock\inst{1} \and Till Riedel\inst{1} \and Likun Fang\inst{1} \and Michael Beigl\inst{1}}
\authorrunning{F. Author et al.}
%
\institute{Karlsruhe Institute of Technology, Karlsruhe, Germany \\ \email{\{yhuang, zhou, hefenbrock, riedel, fang, beigl\}@teco.edu}
}
%
\maketitle              
\begin{abstract}
The vulnerability of the high-performance machine learning models implies a security risk in applications with real-world consequences. Research on adversarial attacks is beneficial in guiding the development of machine learning models on the one hand and finding targeted defenses on the other. However, most of the adversarial attacks today leverage the gradient or logit information from the models to generate adversarial perturbation. Works in the more realistic domain: decision-based attacks, which generate adversarial perturbation solely based on observing the output label of the targeted model, are still relatively rare and mostly use gradient-estimation strategies. In this work, we propose a pixel-wise decision-based attack algorithm that finds a distribution of adversarial perturbation through a reinforcement learning algorithm. We call this method  Decision-based Black-box Attack with Reinforcement learning (DBAR). Experiments show that the proposed approach outperforms state-of-the-art decision-based attacks with a higher attack success rate and greater transferability.

\end{abstract}
\section{Introduction}
Many high-performing machine learning algorithms used in computer vision, speech recognition and other areas are susceptible to minimal changes of their inputs~\cite{brendel2017decision}. Despite their good performance, the vulnerability of machine learning models has raised widespread concerns. Small perturbation on road signs can have a serious impact on automated driving. These actions, which modify the benign input by imperceptible perturbations and thus manipulate the machine learning model to suit the attacker's interests, are called adversarial attacks.

Most adversarial attacks used to construct adversarial perturbation rely either on gradient information (white-box attack) or logit output (score-based attack) of the model. While these approaches help to study the limitations of current machine learning algorithms~\cite{papernot2017practical}, they do not reflect the level of information a real attacker would have access to in most scenarios. In contrast, decision-based attacks consider limited access to the targeted model, allowing only the label information output by the model to be used. Such limited access is far more common in the real-world scenarios making its study more practical.
Most of decision-based attacks start with an adversarial example with large perturbation. Then, adversarial examples with smaller perturbations are gradually found by sample-based gradient estimation. Different attacks exploit the samples in different ways, and therefore the efficiency of the algorithms varies. 
Such gradient-estimation-based approaches, however, require a large number of queries to the targeted model, which affects the efficiency of the algorithm and makes it impossible to perform real-time attack. In addition, the perturbations generated by gradient-based approach are too specific to the particular targeted model and benign example, and therefore lack transferability. To address these shortcomings, we propose a novel pixel-wise decision-based attack approach, called Decision-based Black-box Attack with Reinforcement learning (DBAR), which is guided by rewards instead of gradients. 
We therefore phrase the search for adversarial perturbations as an reinforcement learning task. Depending on whether the learned agent is targeting a single or multiple benign examples, two different attacks are designed, i.e., context-free attack and context-aware attack.



Our contributions can be summarized as: (1) Context-free DBAR achieves state-of-the-art performance, and perturbations sampled from the discovered distribution are more transferable then those generated by the other decision based attacks. In addition, the context-free DBAR is an universal attack which can also attack time-series data and super pixel of image data. (2) The context-aware attack achieves an effective attack without any queries on the targeted model after training, which is not possible for most existing decision-based attacks. (3) The algorithm generates a distribution which can be used to sample multiple different attacks.


\section{Related work}
\label{sec: relatied work}



The definition of decision-based attack was first proposed in~\cite{brendel2017decision}. It starts with an example in the target category and optimizes the attack with random selection and validation. 
This method is simple and effective; however, it is inefficient because the information from the sampled examples is not fully utilized e.g., information from the 'worse' samples. Several methods attempt to bridge this gap. For example, \cite{brunner2019guessing} biases the sampling process by combining low-frequency noise with gradients from surrogate models. However, its performance depends on the transferablity between the surrogate model and the target model.
Similarly, transfer-based attacks~\cite{papernot2017practical} also rely on carefully chosen surrogate models. However they obtain an attack on the original model by attacking the surrogate model. Opt attack~\cite{cheng2018query} transforms the adversarial attack problem into a continuous real-valued optimization problem, i.e., the direction and distance to the decision boundary. This optimization problem can be solved by any zeroth-order optimization algorithm. However, distance calculation and gradient estimation in large dimensions will consume a large number of queries, which reduces the efficiency of the algorithm. 
Evo attack~\cite{dong2019efficient} applies evolutionary algorithms to generate adversarial perturbations and employs some techniques to reduce the dimensionality of the search space. 
It uses a custom variant in normal distribution and update the variant with (1+1)-CMA-ES. However, the variance is sign-independent and the sampling is therefore unstable.
Rays~\cite{chen2020rays} uses the dichotomous method to find perturbations. Although it has achieved good results on many datasets, the effectiveness of the algorithm is difficult to prove as results depend strongly on the test set.
HSJ~\cite{chen2020hopskipjumpattack} estimate gradient in different way and achieve a decision-based attack. However, gradient estimation is time-consuming and, at the same time, reduces the transferability of the generated adversarial perturbations. In this paper, we try to solve the problem without estimating the gradient.



\section{Methodology}
\label{sec: methodology}


We model the decision-based black-box adversarial attack problem as finding the adversarial distribution $p_\Theta$ with parameters $\Theta$ of an $m$-class deep classification model $\mathcal{M}: \mathbb{R}^d \to [m]$ that accepts an input $x \in [0,1]^d$ and outputs $y \in [m] = \left\{1, \cdots, m\right\}$. The objective function can be described as
\begin{equation}
    \min_\Theta\left( \lambda \underset{\eta \sim p_\Theta}{\mathbb{E}}\left\|\eta \right\|_\infty - \underset{\eta \sim p_\Theta}{\mathcal{P}}(\mathcal{M}(x + \eta) \neq \mathcal{M}(x))\right),
    \label{eq: opt_obj}
\end{equation}
where $\eta$ is the adversarial perturbation sampled from the distribution $p_\Theta$, $\left\|\cdot\right\|_\infty$ denotes the $l_\infty$ norm and $\mathcal{P}$ evaluates a probability. The objective function consists of two components, the expected $l_\infty$ norm of the perturbations sampled from the distribution and the attack success rate of the perturbations sampled from the distribution. $\lambda$ is a parameter that trade-off between the expectation and the success rate. The goal of the problem definition is to find a distribution such that the center of the distribution found is as close as possible to the benign example $x$ while the adversarial perturbations sampled from the distribution maintain a high attack success rate.

To solve \eqref{eq: opt_obj} through a reinforcement learning algorithm, depending on whether single or multiple benign examples are considered, we design two different environments: a context-free environment and a context-aware environment, which correspond to context-free attack and context-aware attack.
Both environments share the same setting except the transition model. The state and action space are both set to $\mathbb{R}^d$. The perturbation distribution $p_\Theta$ to optimized is regarded as the agent. Each adversarial perturbation $\eta$ sampled from the distribution is an action and each benign example $x$ is a state. Since the action is continuous, we model the agent with normal distribution $p_\Theta(\eta \mid x)=\mathcal{N}\big(\eta \mid \mu_\Theta(x), \text{diag}\left(\Sigma_\Theta(x)\right)\big)$. A trajectory $\tau$ consists of fixed number of decision step. In each decision step, a perturbation (action) is sample from the distribution (agent) and send to the environment to get the reward and next state.
To achieve the optimization goal, we define the reward function as
\begin{equation}
    r(x, \eta) = \frac{2\cdot \mathbbm{1}_{\left\{\mathcal{M}(x + \eta)\neq \mathcal{M}(x)\right\}}-1}{\left\|\eta\right\|_\infty},
    \label{eq: reward}
\end{equation}
where $\mathbbm{1}$ is the indicator function to identify whether adversarial perturbations mislead the classifier $\mathcal{M}$. When the attack is successful, \mbox{$2\cdot \left(\mathbbm{1}_{\left\{\mathcal{M}(x + \eta)\neq \mathcal{M}(x)\right\}}-1\right) = 1$}, the algorithm can try to increases the reward by shrinking the perturbation. On the other hand, if the attack fails, the algorithm may try to find a successful attack by increasing the perturbation \footnote{Eq.~\ref{eq: reward} can be modified to a target attack by setting the condition of indicator function to $\mathcal{M}(x + \eta) = \text{target}$.}. 
The expectation and probability in \eqref{eq: opt_obj} are approximated by Monte Carlo estimation using sampled trajectories.

In both environment settings, the next state is selected independent of the current state and action. In the context-free environment, only one state, the benign example, exists, while in the context-aware environment, the next state is selected by random sampling.

The objective function can be written as:
\begin{equation}
\begin{split}
    J(\Theta) &= \int p_\Theta(\tau)\left(\sum_{t=0}^Tr(x_t, \eta_t)\right) d\tau,\\
    &p_\Theta(\tau) \approx \prod_{t=0}^{T-1}p_\Theta(\eta_t\vert x_t).
\end{split}
\label{eq: adv_obj}
\end{equation}
To learn $\Theta$, we need to calculate the gradient of the objective function \eqref{eq: adv_obj}. In the black-box adversarial attack, each reward in one trajectory is treated equally and does not depend on the actions in the other time step of the same trajectory. Therefore, when calculating the gradient, at time step $t$, terms that do not depend on the action $\eta_t$ can be omitted. Using the log derivation trick and Monte Carlo sampling~\cite{shapiro2003monte}, the gradient of the objective function can be expressed as

\begin{equation*}
\begin{split}
    \nabla J(\Theta) &\approx \sum_{i=0}^{I} \left[\left(\nabla_\Theta \operatorname{log}\left(\prod_{t=0}^{T-1}p_\Theta(\eta_{t,i}\vert x_{t,i})\right)\right) \left(\sum_{t=0}^T r(x_{t,i}, \eta_{t,i})\right)\right]\\
    &\approx \sum_{i=0}^{I} \left[ \sum_{t=0}^{T-1}r(x_{t,i},\eta_{t,i})\nabla_\Theta \operatorname{log} p_\Theta(\eta_{t,i}\vert x_{t,i})\right]\\
    &= \sum_{i=0}^{I \cdot T}r(x_{i},\eta_{i})\nabla_\Theta\operatorname{log} p_\Theta(\eta_{i}\vert x_{i})
\end{split}
\end{equation*}




So far, this gradient is valid only for the samples generated by $p_\Theta$. We apply the importance sampling 
technique so that old trajectories can be reused. In addition, we limit the update step size as suggested in~\cite{schulman2017proximal}, since importance sampling only works when the update size is small. Together with the stable training trick mentioned in~\cite{mnih2016asynchronous}, the gradient can be expressed as
\begin{equation*}
\begin{split}
    \nabla J(\Theta) 
    &= \sum_{i=0}^{M} \left(\nabla_\Theta\operatorname{min}(w_i(\Theta), \operatorname{clip}(w_i(\Theta), 1-\epsilon, 1+\epsilon))\right)\cdot 
    \left(r(x_i,\eta_i) - V(x_i)\right) \\
    & \text{with} \quad w_i(\Theta) = \frac{p_\Theta(\eta_i\vert x_i)}{p_{\Theta_{\textrm{old}}}(\eta_i\vert x_i)},\\
\end{split}
\end{equation*}

where $p_{\Theta_\textrm{old}}$ is the distribution that generates the training samples and $p_\Theta$ is the distribution, that is frequently updated. The parameter $\epsilon$ limits the update size and $V(x_i)$ is the expected reward given to a benign example $x_i$.

Algorithm \ref{alg: dbar} summarizes the process of generating adversarial distribution, where Actor: $\mathbb{R}^d\to \mathbb{R}^d \times \mathbb{R}^d$ and
Critic: $\mathbb{R}^d\to \mathbb{R}$ are two neural networks with the same ResNet architecture except for the output layer. 


\begin{algorithm}[tb]
\caption{Generating adversarial distribution through DBAR}
\label{alg: dbar}
\textbf{Input}: $x_0$ (benign example), $N$ (number of iterations), $M$ (number of samples in one iteration), $K$ (number of training), $L$ (size of minibatch), $init\_mean$, $init\_std$, $\epsilon$\\
\textbf{Output}: Actor
\begin{algorithmic}[1] 
\STATE Initialization: Initialize Actor($\cdot$) with $init\_mean$ and $init\_std$, Critic($\cdot$), $x \gets x_0$
\FOR{$i \gets 1$ to $N$}
\STATE $B \gets [ \;]$
\FOR{$j \gets 1$ to $M$}
\STATE $\mu$, $\mathbf{I}\cdot \sigma^2$$\gets$ Actor($x$)
\STATE $r, p, x'$ $\gets$ sample action from $\mathcal{N}(\mu,\mathbf{I}\cdot \sigma^2)$, calculate its log-probability and applied it to the environment to get reward and the next benign example
\STATE $B \gets B\cup \left\{\text{r}:r,\text{lp}:lp,\text{x}:x\right\}, x\gets x'$
\ENDFOR
\FOR{$k \gets 1$ to $K$}
\STATE $\left\{B_1,\cdots B_{\left \lfloor M/L \right \rfloor}\right\}$ $\gets$ generate mini-batch from $B$
\FOR{$b \gets \left\{B_1,\cdots B_{\left \lfloor M/L \right \rfloor}\right\}$}
\STATE $\mu', \mathbf{I}\cdot (\sigma')^{2} \gets$ Actor($b[\text{x}]$)
\STATE $lp' \gets$ compute log-probability of $b[\text{x}]$ in $\mathcal{N}(\mu', \mathbf{I}\cdot (\sigma')^2)$
\STATE $v \gets$ Critic($b[\text{x}]$)
\STATE $a \gets b[\text{r}] - v$
\STATE $w \gets \exp(lp'-b[\text{lp}])$
\STATE loss\_actor $\gets \text{min}(w, \text{clip}(w, 1-\epsilon, 1+\epsilon))\cdot a$
\STATE loss\_critic $\gets \text{MSE}(b[\text{r}], v)$
\STATE update Actor with the gradient of loss\_actor
\STATE update Critic with the gradient of loss\_critic
\ENDFOR
\ENDFOR
\ENDFOR
\STATE \textbf{return} Actor
\end{algorithmic}
\end{algorithm}


\section{Experiments}
\label{sec: 4}

In this section, we perform experiments to investigate the following questions:
(i) How does the context-free DBAR algorithm perform on image datasets compared to state-of-the-art decision-based attack methods?
(ii) Can context-free DBAR be applied to time-series datasets?
(iii) Are the perturbations discovered by the context-free DBAR algorithm more transferable than those discovered by state-of-the-art decision-based attack methods?
(iv) Can context-aware DBAR perform real-time attacks after training?
(v) How do the hyper-parameters affect the performance of context-free DBAR?


\subsection{Experiment Setting}
\label{sec: 4.1}

\textbf{Baselines and hyper-parameters}: To evaluate DBAR, we compare it with the following decision-based attacks: (i) the state-of-the-art Decision-based black-box Boundary attack~\cite{brendel2017decision} and the HopSkipJump Attack~\cite{chen2020hopskipjumpattack} for image datasets. (ii) the Universal White-box attack FGSM~\cite{szegedy2013intriguing} and BIM~\cite{kurakin2016adversarial} for time series datasets. 

All attacks are implemented by the python package Foolbox~\cite{rauber2017foolbox}. We use the default hyper-parameter settings for all attacks with a fixed random seed. 
We limit the maximum number of queries for all the attacks to 20000. For the DBAR algorithm, the number of training epoch $K$ is set to $K=10$, with a mini-batch size of $L=10$ and $\epsilon=0.02$. Additionally $init\_mean=0$ and $init\_std=0.5$.

\textbf{Datasets and models}: We carried out attacks over the following image datasets with varied dimensions and dataset sizes: CIFAR10~\cite{krizhevsky2009learning}, CIFAR100~\cite{krizhevsky2009learning}, STL10~\cite{coates2011analysis}, Caltech101~\cite{griffin2007caltech}. The pixel values of all images are normalized to $[0, 1]$. We also attack models with different structures such as ResNet20~\cite{he2016deep} with 272474 parameters and VGG11~\cite{simonyan2014very} with 9756426 parameters. Both models were obtained from Pytorch~\cite{paszke2019pytorch}.
In addition, we carried out attacks over the publicly available time series UCR archiv dataset~\cite{dau2019ucr} and attack the time series ResNet-ts model as defined by~\cite{fawaz2019adversarial}.  

\subsection{Adversarial examples for image and time series data}
\label{sec: 4.2}

We perform non-target attacks on all the image datasets mentioned above and summarise the attack success rate (ASR) of each methods in Table~\ref{tab: exp1}. Concretely, all attacks are applied to $1000$ correctly classified test examples from each dataset. If an adversarial example can mislead the classification model and is in a 0.04 (10/255) $l_\infty$ neighborhood of the benign example, we denote this attack as a success. 
From the result, we see that, context-free DBAR achieves better results on most of the datasets, except for the STL10 dataset on the VGG11 model. Struggling with finding the first success attack is probably the main reason for the failure of the attack. This happens when the $init\_std$ is set too small for the given data set. The influence of the hyper-parameters on the attack performance an run-time is analyzed in experiment~\ref{sec: 4.6}. 

\begin{table}
\centering
\resizebox{\textwidth}{!}{\begin{tabular}{|c|c|c|c|c|c|c|c|c|c|c|c|}
\hline
\multicolumn{2}{|c|}{~}&Model Accuracy&BA&HSJ&DBAR&\multicolumn{2}{|c|}{~}&Model Accuracy&BA&HSJ&DBAR\\
\hline
\multirow{2}*{Cifar10}&ResNet20&0.91&\textbf{1.00}&\textbf{1.00}&\textbf{1.00}&\multirow{2}*{STL10}&ResNet20&0.83&\textbf{1.00}&\textbf{1.00}&\textbf{1.00}\\
\cline{2-6}\cline{8-12}
~&VGG11&0.90&0.71&0.78&\textbf{0.84}&~&VGG11&0.84&0.59&\textbf{0.90}&0.78\\
\hline

\multirow{2}*{Cifar100}&ResNet20&0.66&0.95&0.95&\textbf{0.96}&\multirow{2}*{CalTech101}&ResNet20&0.79&0.93&\textbf{1.00}&\textbf{1.00}\\
\cline{2-6}\cline{8-12}
~&VGG11&0.61&0.63&0.80&\textbf{0.84}&~&VGG11&0.68&0.51&0.80&\textbf{0.82}\\
\hline



\end{tabular}}
\caption{Attack success rate (ASR) of three different decision-based attack methods: Boudary Attack (BA), HopSkipJump Attack (HSJ) and the proposed context-free DBAR, against model ResNet20 and VGG11 on four different image datasets with varies sizes.}
\label{tab: exp1}
\end{table}

%


DBAR is universal in the sense that it is able to attack any example in form of $\mathcal{R}^d$. To prove this, we compare the performance of the proposed context-free DBAR against ResNet-ts model on the UCR open source time series datasets with the two popular  white-box attacks FGSM and BIM. When the size of the perturbation found by an attack is smaller than 0.1, as proposed in~\cite{fawaz2019adversarial}, and the adversarial example can mislead the classification, we regard the attack as success. The parameter settings of this experiment are different from other experiments because the time series data are not normalized to $[0, 1]$. Furthermore, note that DBAR, as opposed to FGSM and BIM is a black-box attack. We remove the step limit and set the initial standard deviation to $80$ to avoid struggling with finding the first successful attack. 


\begin{table}
\centering
\resizebox{\textwidth}{!}{\begin{tabular}{|c|c|c|c|c|c|c|c|c|c|}
\hline
&Model Accuracy&FGSM&BIM&DBAR&&Model Accuracy&FGSM&BIM&DBAR\\
\hline
50words&0.73&0.77&0.88&\textbf{0.91}&DistalAge&0.80&0.78&\textbf{0.79}&0.64\\
\hline

Adiac&0.83&0.96&0.98&\textbf{0.99}&FaceAll&0.86&0.10&0.15&\textbf{0.20}\\
\hline

Beef&0.77&0.74&\textbf{0.87}&\textbf{0.87}&FaceUCR&0.95&0.17&\textbf{0.20}&0.19\\
\hline

Car&0.93&0.76&\textbf{0.92}&0.80&ElectricDevices&0.74&0.34&0.58&\textbf{1.00}\\
\hline

Diatom&0.30&0.00&0.00&\textbf{0.41}&ItalyPowerDemand&0.96&0.04&0.04&\textbf{0.20}\\
\hline

\end{tabular}}
\caption{Attack success rate (ASR) of three different adversarial attack methods: FGSM~\cite{szegedy2013intriguing}, BIM~\cite{kurakin2016adversarial} and the proposed context-free DBAR, against ResNet-ts on ten different time series datasets with varies size and dimensions.}
\label{tab: exp2}
\end{table}

The results can be seen in Table~\ref{tab: exp2}. Although compared to the white-box algorithms, the context-free DBAR achieves better results on six datasets, ties on one dataset, and worse results on three datasets. As for the image datasets, the ASR score on the time series data is independent of the accuracy of the target model. And probably because of the different principles of generating attacks, DBAR performs well on some datasets where BIM performs very poorly, e.g., Diatom, ItalyPowerDemand. At the same time, the opposite situation also exists, see DistalAge. It is important to note that although FGSM has the worst results, it is attacking in real time, while both BIM and the proposed context-free DBAR require multiple iterations. However, a similar real-time attack can be achieved by context-aware DBAR. This is demonstrated in Section~\ref{sec: 4.4}.

\subsection{Transferability of the perturbation distribution and real-time attack}
\label{sec: 4.4}
The high transferability of the perturbation allows attacks generated against one platform to be applied to the other.
To demonstrate the transferability of the perturbation found by the proposed method, we run the experiment in Section~\ref{sec: 4.2} on the Cifar datasets again and apply the attacks generated against the ResNet20 model to the VGG11 model and vice versa. The ASR score is given in Table~\ref{tab: exp3}. 
The performance of the proposed method is significantly better than that of the other methods. In particular, the perturbations generated against the VGG11 model on Cifar100 dataset have a near-average ASR score against the ResNet20 model. Besides, we can see that perturbations generated against simple model (with fewer parameters, ResNet20) are difficult to perform success attack against the more complex VGG11 model. 

\begin{table}
\centering
\resizebox{\textwidth}{!}{\begin{tabular}{|c|c|c|c|c|c|c|c|c|c|}
\hline
&Targeted model&Boundary Attack&HSJ&DBAR&&Targeted model&Boundary Attack&HSJ&DBAR \\
\hline
\multirow{2}*{Cifar10}&ResNet20&0.04&0.04&\textbf{0.13}&\multirow{2}*{Cifar100}&ResNet20&0.14&0.14&\textbf{0.24}\\
\cline{2-5}\cline{7-10}
~&VGG11&0.10&0.10&\textbf{0.36}&~&VGG11&0.14&0.09&\textbf{0.45}\\
\hline


\end{tabular}}
\caption{Attack success rate (ASR) of attacks generated for VGG11 and applied to ResNet20 and vice versa. Targeted model denotes the model used to generate the perturbation (attack).}
\label{tab: exp3}
\end{table}

\begin{figure}[!t]
    \centering
    \includegraphics[width=\textwidth]{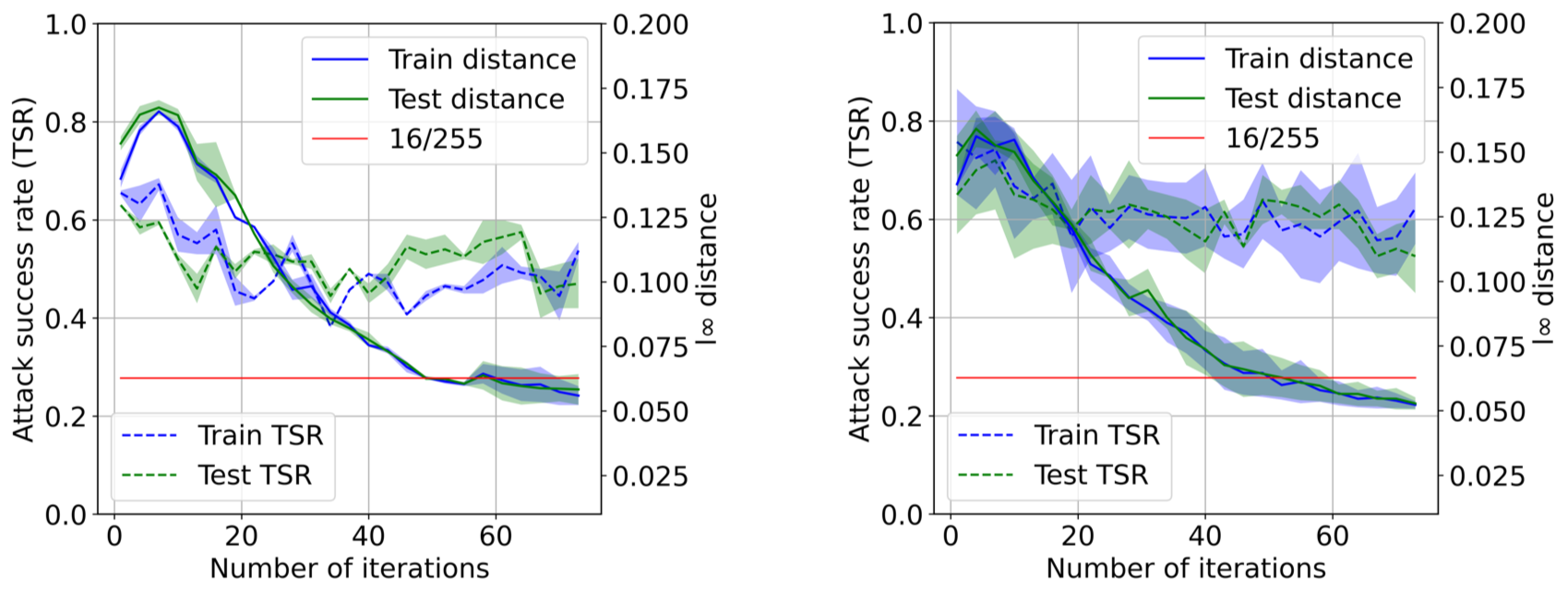}
    \caption{Performance of context-aware DBAR against ResNet20 on Ciar10 (left) and Cifar100 (right) datasets. The picture on the left shows the performance of Cifar10 and the one on the right shows the performance of Cifar100.}
    \label{fig: exp4}
\end{figure}

To evaluate the capabilities for real time attacks, we apply context-aware DBAR against ResNet20 on the Cifar10 and Cifar100. The results can be seen in Fig.~\ref{fig: exp4}. We can find that after 60 iterations, the adversarial perturbation obtained by sampling are able to implement effective attacks that have a success rate of about 50\% on Cifar10 and 60\% on Cifar100. We observe higher $l_\infty$ norms for the perturbations at the beginning, which is most likely caused by the different update directions between the different benign examples. Ideally, the algorithm should reduce the size of the perturbation while increasing the success rate of the training. However, the attack success rate decreases as the training progresses, although the decrease is not significant. The algorithm improves the rewards obtained by reducing the perturbation size. This problem can be mitigated by increasing the contribution of the success attack in the reward function, see~\eqref{eq: reward}.


%

\subsection{Impact of the parameter selection}
\label{sec: 4.6}

\begin{figure*}[!t]
    \centering
    \includegraphics[width=\textwidth]{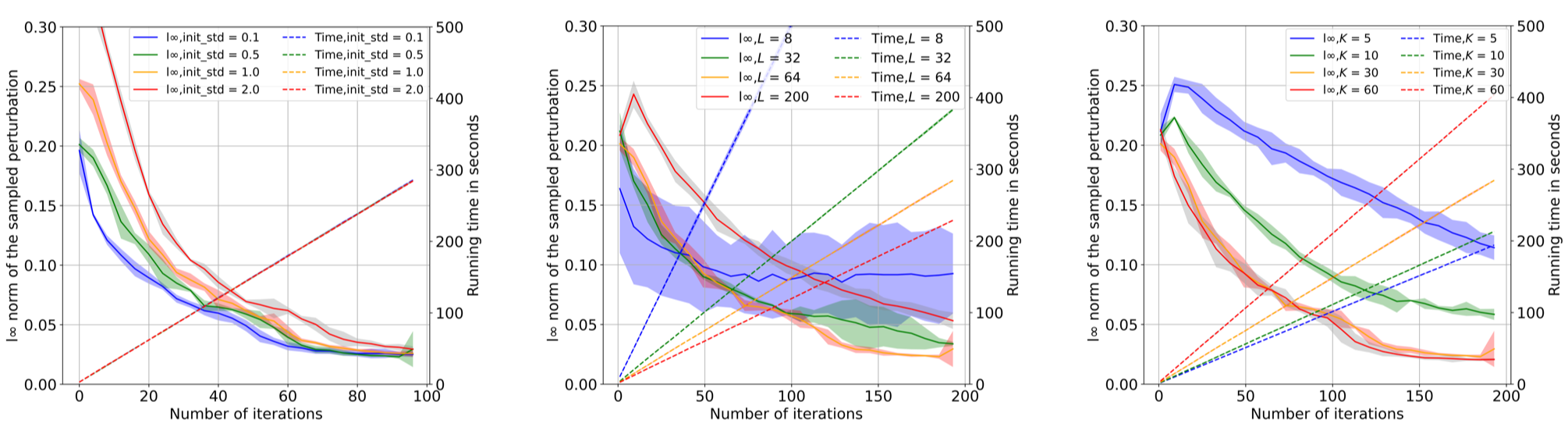}
    \caption{Performance comparison of context-free DBAR with different hyper-parameter settings. The $l_\infty$ norm is shown as a solid line in the plot, while the run time is shown as a dashed line.}
    \label{fig: exp5} 
\end{figure*}
In this experiment, we analyse the effect of following hyper-parameters on the performance of context-free DBAR regarding $init\_std$, number of training epoch in each iteration $K$ and the size of mini batch $L$. The baseline hyper-parameters used in the experiments are set as follows as $init\_std$ = 0.5, $L$ = 64, $K$ = 30. In each trial we modify one of the above parameters and summarise the result in Fig.~\ref{fig: exp5}. Evidently, the parameters have a great impact on the convergence speed, stability and runtime of DBAR. The leftmost plot in Fig.~\ref{fig: exp5} shows the effect of the standard deviation of the initial distribution $init\_std$. The larger the parameter, the larger the perturbation sampled from the initial distribution. In general, the larger the initial perturbation, the higher the probability that it will successfully attack a benign example. When the $init\_std$ is too small and the initial perturbation fails to attack the benign example, the algorithm will struggle in looking for the first successful attack. The size of the training batches $L$ affects the stability, where lower $L$ results in less stability. 

Since the number of samples per iteration is constant, the smaller the batch size, the more times the agent is updated and the longer each iteration takes.

In addition, due to the importance sampling, the log-probability of actions needs to be recalculated before updating agent, which further increases the runtime. These are reflected in the plot in the center of Fig.~\ref{fig: exp5}. The rightmost plot shows that larger number of training $K$ in one iteration converges faster. However, since there are more updates per iteration, it also takes longer to run.

\section{Conclusion}
\label{sec: 5}
In this paper, we formulate the decision-based black-box adversarial attack as a reinforcement learning task and search for the adversarial attack based on reward criterion.
We have experimentally demonstrated the feasibility of the proposed algorithm. In addition, have shown some advantages of the algorithm such as the ability to generate attacks for different kinds of data such as images and time series, the transferability of attacks between different models and the ability for real-time attack. 
Besides, there is still room for further exploration in the proposed approach through using different reward functions or investigating the usage bounds by attacking very small perturbation or high resolution images.

\section{Acknowledgements}
This work was partially funded by the Ministry of The Ministry of Science, Research and the Arts Baden-Wuerttemberg as part of the SDSC-BW and by the German Ministry for Research as well as by Education as part of SDI-C (Grant 01IS19030A)
%
%
%
\bibliographystyle{splncs04}
\bibliography{mybibliography}

\begin{thebibliography}{10}
\providecommand{\url}[1]{\texttt{#1}}
\providecommand{\urlprefix}{URL }
\providecommand{\doi}[1]{https://doi.org/#1}

\bibitem{brendel2017decision}
Brendel, W., Rauber, J., Bethge, M.: Decision-based adversarial attacks:
  Reliable attacks against black-box machine learning models. arXiv preprint
  arXiv:1712.04248  (2017)

\bibitem{brunner2019guessing}
Brunner, T., Diehl, F., Le, M.T., Knoll, A.: Guessing smart: Biased sampling
  for efficient black-box adversarial attacks. In: Proceedings of the IEEE/CVF
  International Conference on Computer Vision. pp. 4958--4966 (2019)

\bibitem{chen2020hopskipjumpattack}
Chen, J., Jordan, M.I., Wainwright, M.J.: Hopskipjumpattack: A query-efficient
  decision-based attack. In: 2020 ieee symposium on security and privacy (sp).
  pp. 1277--1294. IEEE (2020)

\bibitem{chen2020rays}
Chen, J., Gu, Q.: Rays: A ray searching method for hard-label adversarial
  attack. In: Proceedings of the 26th ACM SIGKDD International Conference on
  Knowledge Discovery \& Data Mining. pp. 1739--1747 (2020)

\bibitem{cheng2018query}
Cheng, M., Le, T., Chen, P.Y., Yi, J., Zhang, H., Hsieh, C.J.: Query-efficient
  hard-label black-box attack: An optimization-based approach. arXiv preprint
  arXiv:1807.04457  (2018)

\bibitem{coates2011analysis}
Coates, A., Ng, A., Lee, H.: An analysis of single-layer networks in
  unsupervised feature learning. In: Proceedings of the fourteenth
  international conference on artificial intelligence and statistics. pp.
  215--223. JMLR Workshop and Conference Proceedings (2011)

\bibitem{dau2019ucr}
Dau, H.A., Bagnall, A., Kamgar, K., Yeh, C.C.M., Zhu, Y., Gharghabi, S.,
  Ratanamahatana, C.A., Keogh, E.: The ucr time series archive. IEEE/CAA
  Journal of Automatica Sinica  \textbf{6}(6),  1293--1305 (2019)

\bibitem{dong2019efficient}
Dong, Y., Su, H., Wu, B., Li, Z., Liu, W., Zhang, T., Zhu, J.: Efficient
  decision-based black-box adversarial attacks on face recognition. In:
  Proceedings of the IEEE/CVF Conference on Computer Vision and Pattern
  Recognition. pp. 7714--7722 (2019)

\bibitem{fawaz2019adversarial}
Fawaz, H.I., Forestier, G., Weber, J., Idoumghar, L., Muller, P.A.: Adversarial
  attacks on deep neural networks for time series classification. In: 2019
  International Joint Conference on Neural Networks (IJCNN). pp.~1--8. IEEE
  (2019)

\bibitem{griffin2007caltech}
Griffin, G., Holub, A., Perona, P.: Caltech-256 object category dataset  (2007)

\bibitem{he2016deep}
He, K., Zhang, X., Ren, S., Sun, J.: Deep residual learning for image
  recognition. In: Proceedings of the IEEE conference on computer vision and
  pattern recognition. pp. 770--778 (2016)

\bibitem{krizhevsky2009learning}
Krizhevsky, A., Hinton, G., et~al.: Learning multiple layers of features from
  tiny images  (2009)

\bibitem{kurakin2016adversarial}
Kurakin, A., Goodfellow, I., Bengio, S.: Adversarial machine learning at scale.
  arXiv preprint arXiv:1611.01236  (2016)

\bibitem{mnih2016asynchronous}
Mnih, V., Badia, A.P., Mirza, M., Graves, A., Lillicrap, T., Harley, T.,
  Silver, D., Kavukcuoglu, K.: Asynchronous methods for deep reinforcement
  learning. In: International conference on machine learning. pp. 1928--1937.
  PMLR (2016)

\bibitem{papernot2017practical}
Papernot, N., McDaniel, P., Goodfellow, I., Jha, S., Celik, Z.B., Swami, A.:
  Practical black-box attacks against machine learning. In: Proceedings of the
  2017 ACM on Asia conference on computer and communications security. pp.
  506--519 (2017)

\bibitem{paszke2019pytorch}
Paszke, A., Gross, S., Massa, F., Lerer, A., Bradbury, J., Chanan, G., Killeen,
  T., Lin, Z., Gimelshein, N., Antiga, L., et~al.: Pytorch: An imperative
  style, high-performance deep learning library. Advances in neural information
  processing systems  \textbf{32},  8026--8037 (2019)

\bibitem{rauber2017foolbox}
Rauber, J., Brendel, W., Bethge, M.: Foolbox: A python toolbox to benchmark the
  robustness of machine learning models. arXiv preprint arXiv:1707.04131
  (2017)

\bibitem{schulman2017proximal}
Schulman, J., Wolski, F., Dhariwal, P., Radford, A., Klimov, O.: Proximal
  policy optimization algorithms. arXiv preprint arXiv:1707.06347  (2017)

\bibitem{shapiro2003monte}
Shapiro, A.: Monte carlo sampling methods. Handbooks in operations research and
  management science  \textbf{10},  353--425 (2003)

\bibitem{simonyan2014very}
Simonyan, K., Zisserman, A.: Very deep convolutional networks for large-scale
  image recognition. arXiv preprint arXiv:1409.1556  (2014)

\bibitem{szegedy2013intriguing}
Szegedy, C., Zaremba, W., Sutskever, I., Bruna, J., Erhan, D., Goodfellow, I.,
  Fergus, R.: Intriguing properties of neural networks. arXiv preprint
  arXiv:1312.6199  (2013)

\end{thebibliography}

\end{document}